\documentclass[runningheads]{llncs}
\usepackage{graphicx}
\usepackage{array}
\usepackage{booktabs}

\usepackage[figuresright]{rotating}
\usepackage{xcolor}
\usepackage{float}
\usepackage{longtable}
\usepackage{setspace}
\usepackage{hyperref}
\usepackage{tikz}
\usepackage{caption} 
\usepackage{longtable}

\begin{document}
\title{Error-margin Analysis for Hidden Neuron Activation Labels} 
\titlerunning{Error-margin Analysis for Hidden Neuron Activation Labels}
%
\author{Abhilekha Dalal\inst{1} \and
Rushrukh Rayan\inst{1} \and
Pascal Hitzler\inst{1}}
\authorrunning{Abhilekha Dalal, Rushrukh Rayan, Pascal Hitzler}
%
\institute{Kansas State University, Manhattan, KS, USA} 
\maketitle              
\setlength{\tabcolsep}{6pt}
\begin{abstract}

Understanding how high-level concepts are represented with\-in artificial neural networks is a fundamental challenge in the field of artificial intelligence. 
While existing literature in explainable AI emphasizes the importance of labeling neurons with concepts to understand their functioning, they mostly focus on identifying what stimulus activates a neuron in most cases; this corresponds to the notion of \emph{recall} in information retrieval. We argue that this is only the first-part of a two-part job; it is imperative to also investigate neuron responses to other stimuli, i.e., their \emph{precision}. We call this the neuron label's \emph{error margin}.


\keywords{Explainable AI \and Concept Induction \and CNN}
\end{abstract}

\section{Introduction}

Various explainability AI (XAI) techniques in Deep Learning applications have garnered a lot of traction recently. One of the techniques which has proven its efficacy in explainability is associating high-level human understandable concepts with hidden layer neuron activations \cite{dalal2024value,oikarinen2022clip,tcar_crabbe,tcav_kim,ace_ghorbani}. It is common in the concept-based XAI literature to hand-pick candidate concepts/labels, like the most frequent 20K English words. Previously, we have shown that Concept Induction can be used to assign meaningful labels to neuron activation from a very large knowledge base of candidate concepts used as background knowledge \cite{dalal2024value}\footnote{This is under review at NeSy 2024.} in a scene recognition on images scenario.

Statistical analysis in \cite{dalal2024value} showed that stimuli (network inputs) corresponding to the labels indeed activate the such-labelled neurons with high probability, i.e., the neuron shows high \emph{recall} (in an information retrieval sense) with respect to its "target label". However, the neuron may also activate on many inputs that do not correspond to the neuron's label (e.g., on other neurons' labels, called "non-target labels"), which in information retrieval terms could be understood as low \emph{precision} of the neuron activation, with respect to the assigned label, or in other words, a high false-positives rate, if neuron labels are taken naively at face value.

This is of course not at all unexpected: It is entirely reasonable to assume that any information conveyed by hidden neuron activations be \emph{distributed}, i.e., neurons naturally react to various stimuli, while specific information is indicated by simultaneous activation of neuron groups. However, in order for neuron labelling as in \cite{dalal2024value} to be practically useful, one would like to use hidden neuron activations to "read off" what the network has detected wrt.{} a stimulus, i.e. a high false-positive rate is problematic in this scenario.

Herein, we address this issue as follows. We show that the analysis in \cite{dalal2024value} makes it possible to assign \emph{error margins} to neuron target labels. If a neuron is activated by a stimulus, then the error margin indicates the likelihood that the stimulus indeed falls under the neuron's target label, and this likelihood can be conveyed to the user. We statistically validate error margins by means of data obtained from a user experiment conducted on Amazon Mechanical Turk. 

\section{Method}
\label{sec:method}


In this section, we outline our technical approach for assessing neuron-label associations through \textit{error-margin} analysis (Non-target Label Activation Percentage, or Non-TLA). Non-TLA represents the percentage of images \emph{not falling under the target label} that activate a neuron that carries the target label as per the prior analysis. Similarly, Target Label Activation Percentage, TLA, represents the percentage of images falling under the target label that activate the neuron that carries the target label. 

To obtain error margins, we calculate activation percentages for both target labels and non-target labels per neuron based on Google Images retrieved from the labels as search terms, and we also take into account activation patterns of neuron groups for semantically related labels, analyzing TLA and Non-TLA across different cutoff values. We then use images from the ADE20K dataset~\cite{zhou2019semantic}, with annotations improved thorugh Amazon Mechanical Turk, to statistically validate the error-margins obtained earlier. 
%
The experimental setting is the same as \cite{dalal2024value} 
which we briefly outline below.

\smallskip\noindent{\textbf{Background Premise}}
\label{par:bakgroundpremise}
The primary objective of \cite{dalal2024value} is to provide insights into the contributions of hidden layer neurons within a Convolutional Neural Network (CNN) in  classification tasks. We explore the task of Scene classification using ResNet50V2 on the ADE20K dataset. After training, 
we examine the activation patterns of the last hidden layer neurons and assign high-level concepts to each neuron, by making use of 
OWL-reasoning-based Concept Induction over a background knowledge base derived from Wikipedia, comprising approximately 2 million classes, e.g., neuron 1 gets assigned "cross walk" as label. A statistical analysis in \cite{dalal2024value} shows that each neuron that gets a label assigned (called the neuron's \emph{target label}) indeed activates particularly on network inputs that fall under this label (e.g., images showing a cross walk, for neuron 1): The neuron as an indicator of the target label concept has high recall. However, as the data in \cite{dalal2024value} also indicates, neurons also tend to activate for many other inputs, i.e. they tend to have relatively low precision. 

\smallskip\noindent{\textbf{Computation of Non-TLA}} Concept Induction in fact generates a number of concept labels for each neuron unit, ranked by some accuracy measure. Herein, we consider the Top-3 labels (ranked by coverage score) for each of the 64 neurons in the dense layer. Using the Target-Label image dataset (each image falls under the target label), the TLA 
is calculated, and, using a Non-target Label image dataset (none of the images contain the target label), the Non-TLA is calculated, for each neuron at specified activation value thresholds, namely $> 0$, $> 20\%$, $> 40\%$, and $> 60\%$ of the max activation value that was recorded for each neuron. E.g., (see Table \ref{tab:listofConcepts_google}), neuron 43 activates at $> 40\%$ of its max activation value in about 19.7\% of images \emph{not} showing a central reservation.



\smallskip\noindent{\textbf{Neuron Ensembles for Concept Associations}} The distribution of input information across simultaneously activated neurons necessitates the investigation of neuron ensemble activations at different cut-off activation values. However, an exhaustive analysis of all neuron ensembles does not scale as even just 64 neurons give rise to $2^{64}$ possible neuron ensembles. We deal with this by considering only ensembles of neurons that activate for semantically related labels. 
For example, the concept \textit{building} activates both neurons 0 and 63 (see Table~\ref{tab:listofConcepts_google}); we evaluate all images from Non-target Label image dataset as well as Target Label image dataset \textit{separately}, activating neurons 0 and 63 at the specified cut-off activation values, to calculate TLA and Non-TLA. 
In scenarios where a concept activates more than two neurons, our analysis encompasses all possible combinations of pairs, triples, etc., of neurons (see \emph{skyscraper} in Table ~\ref{tab:listofConcepts_google}). 
We then narrow our focus to a list of highly associated concepts corresponding to the neurons (see the Concepts column in Table~\ref{tab:listofConcepts_google}), that demonstrate TLA exceeding $80\%$, i.e., those neurons with high recall. 

\begin{table}[tb]
\caption{Selective representation of Non-TLA for Google Image dataset(full version can be found in Appendix~\ref{tab:appendix_listofConcepts_google}): The table showcases a refined selection, inclusive of concepts with TLA $>$ 80. Non-t: percentage of non-target label images that activate the neuron(s) associated with the concept being analyzed across various activation thresholds.}
\label{tab:listofConcepts_google}
\centering
\resizebox{\columnwidth}{!}{
\begin{footnotesize}
\begin{tabular}{lr|r|rr|rr}
\hline
Concepts & Neuron & \textbf{targ \%$>$0} & \multicolumn{4}{c}{Non-target \% for different threshold values} \\
\hline
\multicolumn{2}{c}{} &  & non-t $>$0 & non-t $>$ 20\% & non-t $>$ 40\% & non-t $>$ 60\%\\
\hline
buffet & 62 & \textbf{83.607} & 32.714 & 12.374 & 3.708 & 0.825 \\
building & 0 & \textbf{89.024} & 72.328 & 39.552 & 12.040 & 2.276 \\
building & 0, 63 & \textbf{80.164} & 43.375 & 12.314 & 2.276 & 0.182 \\
building and dome & 0 & \textbf{90.400} & 78.185 & 45.133 & 14.643 & 2.639 \\
central\_reservation & 43 & \textbf{95.541} & 84.973 & 57.993 & 19.734 & 2.913 \\
\hline
tap and shower\_screen & 36 & \textbf{86.250} & 72.584 & 32.574 & 7.836 & 0.860 \\
teapot and saucepan & 30 & \textbf{81.481} & 47.984 & 18.577 & 4.367 & 0.845 \\
wardrobe and air\_conditioning & 19 & \textbf{89.091} & 65.034 & 31.795 & 6.958 & 1.145\\
skyscraper & 22 & \textbf{99.359} & 54.893 & 21.914 & 0.977 & 0.977 \\
skyscraper & 54 & \textbf{98.718} & 70.432 & 26.851 & 7.050 & 0.941 \\
\hline
skyscraper & 63 & \textbf{94.393} & 51.612 & 20.618 & 5.775 & 1.143 \\
skyscraper & 22, 54 & \textbf{97.165} & 47.422 & 7.910 & 0.465 & 0.000 \\
skyscraper & 22, 63 & \textbf{96.947} & 36.408 & 5.521 & 0.449 & 0.008 \\
\hline
skyscraper & 54, 63 & \textbf{96.074} & 37.149 & 5.594 & 0.615 & 0.046 \\
skyscraper & 22, 54, 63 & \textbf{95.420} & 29.090 & 3.023 & 0.234 & 0.000 \\
skyscraper & 26, 54, 63 & \textbf{81.134} & 16.823 & 1.975 & 0.350 & 0.023 \\
skyscraper & 22, 26, 54, 63 & \textbf{80.589} & 13.093 & 0.872 & 0.015 & 0.000 \\
\hline
\end{tabular}
\end{footnotesize}}
\end{table}

\smallskip\noindent{\textbf{Annotations of ADE20K Dataset}} 
The analysis just described yields \textit{error-margins} associated with each concept, for each of the chosen activation thresholds listed in Table \ref{tab:listofConcepts_google}. 
For example, the concept \textit{buffet} has an \textit{error-margin} of 12.374 for the Non-TLA of $> 20\%$: Our analysis suggests the \emph{hypothesis} that at most 12.374\% of \textit{non-buffet} images activate the neuron unit 62 at $20\%$ of its max activation value. In other words, the \textit{error-margin} at Non-TLA of $> 20\%$ for the concept \textit{buffet} is 12.374\%. If this hypothesis can be substantiated, then upon presentation of a new input to the network, activation of neuron 62 of at least $20\%$ of its max activation value means that a \emph{buffet} has been detected, and that this detection is \emph{wrong} in at most about 12.374\% of cases.

In order to substantiate our hypotheses, we analyse neuron activation values for new inputs, more precisely for images taken from the ADE20K dataset that was also used in \cite{dalal2024value}. We take advantage of the fact that ADE20K images already carry rich object annotations, however we have observed that they are still too incomplete for our purpuses. Therefore we made use of Amazon Mechanical Turk via the Cloud Research platform, to add missing annotations from a list of concepts derived from Table\ref{tab:listofConcepts_google} to 1050 randomly chosen ADE20K images. Details of the study are in Appendix~\ref{subsec:appendix_amt}.

\smallskip\noindent{\textbf{Validating Neuron-Concept Associations}} 
%
%
To assess the validity of the \textit{error-margins} retrieved from the Google Image dataset for all concepts in Table~\ref{tab:listofConcepts_google}, we look at activations yielded by ADE20K images, and hypothesize that they are similar or lower (i.e., not higher), for non-target images. 
Non-TLA are computed across the predefined cut-off activation thresholds. 
Selected values can be found in Table ~\ref{tab:non_targ_ade_google}.
%
%
For example, the central reservation neuron 43 mentioned above activates above its 40\%{} max activation threshold for about 14.9\%{} of ADE20K non-target images (not showing central reservations), while it activates for about 19.7\%{} of Google non-target images.
%
Both single-neuron and neuron ensemble activations are considered and shown in Table~\ref{tab:non_targ_ade_google}. 

\begin{table}[tb]
\caption{Selective representation of Non-TLA for ADE20K and Google Image dataset (for full version see Appendix~\ref{tab:appendix_non_targ_ade_google}): Non-t: percentage of non-target label images that activate the neuron(s) associated with the concept being analyzed across various activation thresholds.}
\label{tab:non_targ_ade_google}
\centering
\resizebox{\columnwidth}{!}{
\begin{footnotesize}
\begin{tabular}{l|rr|rr|rr|rr}
\hline
Concepts & \multicolumn{2}{c}{non-t $>$0} & \multicolumn{2}{c}{non-t $>$20\%} & \multicolumn{2}{c}{non-t $>$40\%} & \multicolumn{2}{c}{non-t $>$60\%}\\
\hline
 & Google & ADE20K & Google & ADE20K & Google & ADE20K & Google & ADE20K\\
\hline
buffet & 32.714 & 40.135 & 12.374 & 25.817 & 3.708 & 9.470 & 0.825 & 1.804 \\
building & 43.375 & 11.458 & 12.314 & 5.208 & 2.276 & 1.458 & 0.182 & 0.000 \\
building and dome & 78.185 & 26.170 & 45.133 & 5.893 & 14.643 & 0.867 & 2.639 & 0.000 \\
central\_reservation & 84.973 & 44.893 & 57.993 & 34.343 & 19.734 & 14.927 & 2.913 & 3.816 \\
clamp\_lamp and clamp & 59.504 & 27.273 & 29.229 & 19.170 & 9.000 & 8.300 & 1.652 & 1.976 \\
closet and air\_conditioning & 71.054 & 30.168 & 38.491 & 15.620 & 10.135 & 5.513 & 1.267 & 1.378 \\
cross\_walk & 28.241 & 21.474 & 6.800 & 16.391 & 1.524 & 9.784 & 0.521 & 2.922 \\
edifice and skyscraper & 48.761 & 24.187 & 21.786 & 8.453 & 8.379 & 1.300 & 2.229 & 0.260 \\
faucet and flusher & 78.562 & 56.967 & 37.862 & 30.580 & 12.104 & 11.097 & 1.873 & 1.850 \\

\hline
\end{tabular}
\end{footnotesize}}
\end{table}

\section{Statistical Evaluation and Results}
\label{sec:eval}
For a statistical evaluation of our error margin values, we treat each row, representing a concept-error pair at each threshold level, from Table~\ref{tab:non_targ_ade_google}, as an individual hypothesis. For example, the \textit{error-margin} (Non-TLA) for the concept "central reservation" under the $>$ 40 threshold constitutes one hypothesis. 
This way, we get $33 \times 4 = 132$ hypotheses to test. 

We conduct Mann-Whitney U tests (MWU)~\cite{McKnight2010MannWhitneyUT} with the null hypothesis (H0) stating that there is no difference in Non-TLA across both datasets, while the alternative hypothesis (H1) posits that Non-TLA in Google Images is greater than in the ADE20K dataset. 
We choose the MWU test for its robustness with non-parametric data and its aptitude for comparing distributions of independent samples. As our Non-TLA data may not adhere to normality and we're comparing distinct datasets (Google Images and ADE20K), the MWU test provides a reliable means to analyze differences in Non-TLA.


Table~\ref{tab:stats_eval} presents a comparison of Non-TLA between the Google Images and ADE20K datasets for all concepts. Each row represents a concept, with columns displaying the percentage of non-target label images activating associated neuron(s) in both datasets. The p-values from the MWU test indicate the significance of differences in Non-TLA between the datasets. The analysis reveals a consistent trend of decreased Non-TLA in the ADE20K dataset compared to Google Images across various threshold categories. Among the 33 hypotheses tested for the category of Non-TLA $>$ 0, 13 were rejected at a significance level of p $<$ 0.05. Similarly, for Non-TLA $>$ 20\%, 15 hypotheses were rejected at the same significance level. In the case of Non-TLA $>$ 40\%, 21 hypotheses were rejected, while for Non-TLA $>$ 60\%, 23 hypotheses were rejected, all at a p-value $<$ 0.05. Concepts with p-value $<$ 0.05 are deemed statistically significant and are identified as \textit{confirmed} concepts, subject to further scrutiny for their reliability and potential implications.

\begin{table}[tb]
\caption{Selective representation of Statistical Evaluation for \textit{confirmed} concepts - getting $p$-value $<$0.05 for MWU (full version can be found at Appendix~\ref{tab:appendix_stats_eval}: G: activations for Google Image dataset; A: activations for ADE20K dataset; Non-t: percentage of non-target images activating the associated neuron(s) analyzed across various activation thresholds.}
\label{tab:stats_eval}
\centering
\resizebox{\columnwidth}{!}{
\begin{footnotesize}
\begin{tabular}{l|rr|r}
\hline
Concepts & G & A & $p$\\
\hline
\hline
 & \multicolumn{2}{c}{non-t $>$0} & \\
\hline
building & 43.4 & 11.5 & $<$0.05 \\
building and dome & 78.2 & 26.2 & $<$0.05 \\
central\_reservation & 84.9 & 44.8 & $<$0.05 \\
\hline
\multicolumn{3}{c}{\textbf{Wilcoxon signed rank test (non-t $>$0)}} & \textbf{$<$0.05}\\
\hline
\hline & \multicolumn{2}{c}{non-t $>$20 \%} & \\
\hline
 building & 12.3 & 5.2 & $<$0.05 \\
building and dome & 45.2 & 5.8 & $<$0.05 \\
clamp\_lamp and clamp & 29.3 & 19.2 & $<$0.05 \\
closet and air\_conditioning & 38.5 & 15.6 & $<$0.05 \\
\hline
\multicolumn{3}{c}{\textbf{Wilcoxon signed rank test (non-t $>$ 20\%)}} & \textbf{$<$0.05}\\
\hline
\end{tabular}

\begin{tabular}{l|rr|r}
\hline
Concepts & G & A & $p$\\
\hline
\hline
 & \multicolumn{2}{c}{non-t $>$40 \%} & \\
\hline
building & 2.3 & 1.4 & $<$0.05 \\
building and dome & 14.6 & 0.8 & $<$0.05 \\
central\_reservation & 19.7 & 14.9 & $<$0.05 \\
clamp\_lamp and clamp & 9.1 & 8.3 & $<$0.05 \\
\hline
\multicolumn{3}{c}{\textbf{Wilcoxon signed rank test (non-t $>$ 40\%)}} & \textbf{$<$0.05}\\
\hline
\hline
 & \multicolumn{2}{c}{non-t $>$60 \%} & \\
\hline
building & 0.1 & 0.0 & $<$0.05 \\
building and dome & 2.6 & 0.0 & $<$0.05 \\
central\_reservation & 2.9 & 3.8 & $<$0.05 \\
\hline
\multicolumn{3}{c}{\textbf{Wilcoxon signed rank test (non-t $>$ 60\%)}} & \textbf{$<$0.05}\\
\hline
\end{tabular}
\end{footnotesize}}
\end{table}

After confirming concepts using the MWU, we proceed to validate them further using Wilcoxon signed-rank tests. To calculate the Wilcoxon test, we used an online website calculator 
available at \footnote{\url{http://www.statskingdom.com/170median_mann_whitney.html}}.
We employ the Wilcoxon test, with the hypothesis that the difference between Non-TLA of ADE20K and Google Image dataset would be less than or equal to zero (H0), while the alternative hypothesis (H1) suggested a decrease in Non-TLA in the ADE20K dataset compared to the Google image dataset. 
Each threshold serves as an individual hypothesis for the Wilcoxon test, with Non-TLA of the \textit{confirmed} concepts for Google and ADE20K datasets grouped accordingly. For instance, all confirmed Non-TLA $>$ 0 for both datasets constitute one hypothesis, while those $>$ 20\% form another. The p-values, denoting the significance of the test results, are displayed at the bottom of the table. Remarkably, the obtained p-values for each threshold suggest the rejection of the null hypothesis, indicating statistically significant differences in Non-TLA between the datasets when considered separately.
A p-value $<$ 0.05 from this test would indicate a statistically significant decrease in Non-TLA in the ADE20K dataset compared to the Google dataset, further strengthening our findings and highlighting that the error estimates from the Google image data hold, or are even bettered by, the ADE20K images.

We also examine all \textit{confirmed} concepts from all thresholds together in the Wilcoxon test with the same alternative hypothesis ((H1) suggested a decrease in Non-TLA in the ADE20K dataset compared to the Google image dataset), which provides a comprehensive overview of the differences in Non-TLA between the Google and ADE20K datasets across various levels of activation thresholds. This approach aggregates the results from individual thresholds, offering a more consolidated perspective on the overall significance of the differences observed. In our analysis, obtaining a p-value of 5.633e-7, which is less than 0.05, implies the rejection of the null hypothesis. This indicates a statistically significant decrease in Non-TLA in the ADE20K dataset compared to the Google Image dataset when considering all thresholds collectively.
\section{Conclusion}
\label{sec:conclusion}



In this study, we addressed the challenge of understanding high-level concepts within neural networks by proposing a methodology to label neurons with concepts, thereby enhancing model interpretability in explainable AI. Our approach goes beyond identifying activating stimuli for neurons by examining their responses to both intended and unintended concepts. Through systematic analysis and empirical validation using datasets like Google Images and ADE20K, we have demonstrated the effectiveness and generalizability of our method. Statistical evaluation confirms the reliability of error margins obtained from Google Images. Notably, we observed consistent trends indicating decreased non-target activations in the ADE20K dataset, highlighting its potential for robust image analysis tasks. Our contributions include insights into assigning reliable labels to hidden neuron responses, offering a systematic approach to analyze neuron responses to target and non-target concepts, and enhancing the transparency and reliability of concept-based explainable AI systems. 

\paragraph{Acknowledgement.} The authors acknowledge partial funding under National Science Foundation grant 2119753.


\bibliographystyle{splncs04}
\bibliography{reference}
\newpage
\appendix
\section{Appendices}
\label{appendix}

\subsection{Related work}
\label{sec:literature_review}

Efforts to demystify deep learning~\cite{gunning2019xai,adadi2018peeking,minh2022explainable} are ongoing. Methods for explainability can be categorized based on their approach to understanding input data, such as feature summarization~\cite{selvaraju2016grad,lime_rebeiro}, or the model's internal representation, like node summarization~\cite{zhou2018interpreting,bau2020understanding}. These methods further classify into model-specific~\cite{selvaraju2016grad} or model-agnostic~\cite{lime_rebeiro} approaches. Some methods rely on human interpretation of explanatory data, such as posing counterfactual questions~\cite{DBLP:journals/corr/abs-1711-00399}.

Model-agnostic techniques for feature attribution, such as LIME~\cite{lime_rebeiro} and SHAP~\cite{shap_lundberg}, aim to elucidate model predictions by assessing the influence of individual features. However, they encounter challenges like explanation instability~\cite{attacklime_alvarez} and susceptibility to biased classifiers~\cite{attacklimeandshap_lakkaraju}. On the other hand, pixel attribution endeavors to comprehend predictions by assigning significance to individual pixels~\cite{saliencymap_simonyan,gradcam_batra,smoothgrad_smilkov}. Nonetheless, it faces notable limitations, particularly with ReLU activation~\cite{gradcamattack_shrikumar} and adversarial perturbations~\cite{unreliablepixelattribution_kim}, leading to inconsistencies in interpretability.

Explanations developed by \cite{tcav_kim,tcar_crabbe} employ supervised learning and curated concepts. These methods utilize classifiers on target concepts, with weights representing Concept Activation Vectors (CAVs). Another approach by \cite{ace_ghorbani} utilizes image segmentation and clustering for concept selection, albeit potentially losing information and only applicable to visible concepts. \cite{invertibletcav_zhang} proposed enhancements using Non-negative Matrix Factorization to mitigate information loss. Individual Conditional Expectation (ICE) plots~\cite{ice_goldstein} and Partial Dependency Plots~\cite{pdp_friedman} provide insights into prediction-feature relationships from both local and global perspectives but may struggle with intricate feature interactions.

Previous studies suggest that hidden neurons may represent high-level concepts~\cite{zhou2018interpreting,bau2020understanding}, but these methods often require semantic segmentation~\cite{xiao2018unified} (resource-intensive) or explicit concept annotations~\cite{tcav_kim}. 
Some research have utilized Semantic Web data for explaining deep learning models~\cite{confalonieri2021using,diaz2022explainable}, and Concept Induction for providing explanations~\cite{sarker2017explaining,9736291}. However, their focus was on analyzing input-output behavior, generating explanations for the overall system. 

CLIP-Dissect~\cite{oikarinen2022clip} represents work similar to ours, employing a different approach. They utilize the CLIP pre-trained model, employing zero-shot learning to associate images with labels. Label-Free Concept Bottleneck Models~\cite{oikarinen2023label}, building upon CLIP-Dissect, use GPT-4~\cite{achiam2023gpt} for concept set generation.
However, CLIP-Dissect has limitations that may be challenging to overcome without significant changes to the approach. These include limited accuracy in predicting output labels based on concepts in the last hidden layer and difficulty in transferring to other modalities or domain-specific applications. The Label-Free approach inherits these limitations and may compromise explainability, as it uses a concept derivation method that is not inherently explainable.

\subsection{Details of AMT user-study}
\label{subsec:appendix_amt}
Using a subset of randomly chosen 1050 ADE20K images, we conducted a user study through Amazon Mechanical Turk using the Cloud Research platform, to annotate images based on a list of concepts derived from Table\ref{tab:appendix_listofConcepts_google}. 

The study protocol was reviewed and approved by the Institutional Review Board (IRB) at Kansas State University and was deemed exempt under the criteria outlined in the Federal Policy for the Protection of Human Subjects, 45 CFR §104(d), category: Exempt Category 2 Subsection ii. The study was conducted in 35 batches (each batch containing 30 images), with 5 participants per study compensated with \$5 for completing the task. The task was estimated to take approximately 40 minutes, equivalent to \$7.50 per hour. 

For each image, users were presented with a list of concepts (a concise form of concepts from 
Table~\ref{tab:appendix_listofConcepts_google}) to choose from, including buffet, building, building and dome, central\_reservation, clamp\_lamp and clamp, closet and air-conditioning, cross\_walk, edifice and skyscraper, faucet and flusher, field, flusher and soap\_dish, footboard and chain, hedgerow and hedge, lid and soap\_dispenser, mountain, mountain and bushes, night\_table, open\_fireplace and coffee\_table, pillow, potty and flusher, road, road and automobile, road and car, route, route and car, shower\_stall and cistern, Shower\_stall and screen\_door, skyscraper, slope, tap and crapper, tap and shower\_screen, teapot and saucepan, wardrobe and air-conditioning. 

Users were allowed to select multiple concepts for each image, indicating all concepts that applied to the given image. These selected concepts were considered annotations for the respective image.

\subsection{Detailed result of Non-TLA  and Statistical Evaluation}
The detailed results of Non-target Label Activation Percentages (Non-TLA) for the Google dataset are meticulously outlined in Table \ref{tab:appendix_listofConcepts_google}. This table presents a carefully curated selection, focusing on concepts and neuron ensembles with Target Label Activation (TLA) exceeding 80\%. It offers valuable insights into the percentage of Non-Target Label images activating the neuron(s) associated with the concept under scrutiny across a spectrum of activation thresholds.

Furthermore, Table \ref{tab:appendix_non_targ_ade_google} compares the Non-TLA between the ADE20K and Google Image datasets, highlighting variations in activation across different thresholds and providing a comprehensive view of dataset-specific nuances.

Lastly, the statistical evaluation for \textit{confirmed} concepts, outlined in Table \ref{tab:appendix_stats_eval}, underscores concepts where statistical significance (with a $p$-value less than 0.05 for the MWU test) has been established. This table offers insights into the percentage of non-target label images activating the associated neuron(s) across diverse activation thresholds, providing valuable information on the robustness of the identified concepts across both the Google and ADE20K datasets.

\begin{table}[p]
\caption{Non-target Label Activation Percentages (Non-TLA) for Google dataset: The table showcases a refined selection, inclusive of concepts and neuron ensembles with targ(et) activation $>$ 80\%. Non-t: percentage of non-target images that activate the neuron(s) associated with the concept being analyzed across various activation thresholds.}
\label{tab:appendix_listofConcepts_google}
\centering
\resizebox{\columnwidth}{!}{
\begin{footnotesize}
\begin{tabular}{lr|r|rr|rr}
\hline
Concepts & Neuron & \textbf{targ \%$>$0} & \multicolumn{4}{c}{Non-target \% for different threshold values} \\
\hline
\multicolumn{2}{c}{} &  & non-t $>$0 & non-t $>$ 20\% & non-t $>$ 40\% & non-t $>$ 60\%\\
\hline
buffet & 62 & \textbf{83.607} & 32.714 & 12.374 & 3.708 & 0.825 \\
building & 0 & \textbf{89.024} & 72.328 & 39.552 & 12.040 & 2.276 \\
building & 0, 63 & \textbf{80.164} & 43.375 & 12.314 & 2.276 & 0.182 \\
building and dome & 0 & \textbf{90.400} & 78.185 & 45.133 & 14.643 & 2.639 \\
central\_reservation & 43 & \textbf{95.541} & 84.973 & 57.993 & 19.734 & 2.913 \\
\hline
clamp\_lamp and clamp & 7 & \textbf{95.139} & 59.504 & 29.229 & 9.000 & 1.652 \\
closet and air\_conditioning & 19 & \textbf{86.891} & 71.054 & 38.491 & 10.135 & 1.267 \\
cross\_walk & 1 & \textbf{88.770} & 28.241 & 6.800 & 1.524 & 0.521 \\
edifice and skyscraper & 63 & \textbf{92.135} & 48.761 & 21.786 & 8.379 & 2.229 \\
faucet and flusher & 29 & \textbf{95.695} & 78.562 & 37.862 & 12.104 & 1.873 \\
\hline
field & 18 & \textbf{91.824} & 65.333 & 30.207 & 8.183 & 1.656 \\
flusher and soap\_dish & 56 & \textbf{90.094} & 63.552 & 29.901 & 7.695 & 1.148 \\
footboard and chain & 49 & \textbf{88.889} & 66.702 & 40.399 & 17.064 & 4.399 \\
hedgerow and hedge & 54 & \textbf{91.165} & 68.527 & 30.421 & 7.685 & 1.352 \\
lid and soap\_dispenser & 29 & \textbf{99.237} & 78.571 & 34.989 & 9.052 & 1.485 \\
\hline
mountain and bushes & 16 & \textbf{87.037} & 24.969 & 10.424 & 4.666 & 1.937 \\
mountain and bush & 16 & \textbf{87.037} & 24.969 & 10.424 & 4.666 & 1.937 \\
mountain & 43 & \textbf{99.367} & 88.516 & 64.169 & 23.112 & 4.326 \\
night\_table & 3 & \textbf{90.446} & 56.714 & 27.691 & 7.691 & 1.137 \\
open\_fireplace and coffee\_table & 41 & \textbf{88.525} & 16.381 & 4.325 & 0.812 & 0.088 \\
\hline
pillow & 3 & \textbf{98.214} & 61.250 & 28.228 & 7.249 & 1.001 \\
pillow & 50 & \textbf{99.405} & 66.834 & 24.242 & 4.101 & 0.530 \\
pillow & 3, 50 & \textbf{97.605} & 46.492 & 9.634 & 0.988 & 0.049 \\
potty and flusher & 29 & \textbf{88.525} & 76.830 & 36.537 & 10.755 & 1.932 \\
road and car & 51 & \textbf{98.810} & 48.571 & 25.373 & 8.399 & 3.261 \\
\hline
road and automobile & 51 & \textbf{92.560} & 41.466 & 16.055 & 3.301 & 0.701 \\
road & 48 & \textbf{100.000} & 76.789 & 47.897 & 18.843 & 3.803 \\
road & 48, 51 & \textbf{97.099} & 44.592 & 17.727 & 3.471 & 0.702 \\
route & 48 & \textbf{100.000} & 80.834 & 51.873 & 21.034 & 4.979 \\
route and car & 51 & \textbf{92.628} & 47.408 & 18.871 & 4.081 & 1.416 \\
\hline
route & 48, 51 & \textbf{94.334} & 45.089 & 18.937 & 4.809 & 1.169 \\
shower\_stall and cistern & 8 & \textbf{100.000} & 53.186 & 24.788 & 8.485 & 1.372 \\
Shower\_stall and screen\_door & 57 & \textbf{98.496} & 31.747 & 12.876 & 4.121 & 1.026 \\
slope & 18 & \textbf{92.143} & 64.503 & 29.976 & 6.894 & 1.200 \\
tap and crapper & 36 & \textbf{89.130} & 70.606 & 36.839 & 13.696 & 2.511 \\
\hline
tap and shower\_screen & 36 & \textbf{86.250} & 72.584 & 32.574 & 7.836 & 0.860 \\
teapot and saucepan & 30 & \textbf{81.481} & 47.984 & 18.577 & 4.367 & 0.845 \\
wardrobe and air\_conditioning & 19 & \textbf{89.091} & 65.034 & 31.795 & 6.958 & 1.145\\
skyscraper & 22 & \textbf{99.359} & 54.893 & 21.914 & 0.977 & 0.977 \\
skyscraper & 54 & \textbf{98.718} & 70.432 & 26.851 & 7.050 & 0.941 \\
\hline
skyscraper & 63 & \textbf{94.393} & 51.612 & 20.618 & 5.775 & 1.143 \\
skyscraper & 22, 26 & \textbf{82.116} & 22.274 & 3.423 & 0.292 & 0.004 \\
skyscraper & 26, 54 & \textbf{82.225} & 28.782 & 5.444 & 0.703 & 0.054 \\
skyscraper & 22, 54 & \textbf{97.165} & 47.422 & 7.910 & 0.465 & 0.000 \\
skyscraper & 22, 63 & \textbf{96.947} & 36.408 & 5.521 & 0.449 & 0.008 \\
\hline
skyscraper & 26, 63 & \textbf{81.788} & 21.421 & 3.335 & 0.534 & 0.088 \\
skyscraper & 54, 63 & \textbf{96.074} & 37.149 & 5.594 & 0.615 & 0.046 \\
skyscraper & 22, 26, 54 & \textbf{81.461} & 18.940 & 2.363 & 0.169 & 0.000 \\
skyscraper & 22, 26, 63 & \textbf{81.243} & 15.252 & 1.706 & 0.184 & 0.004 \\
skyscraper & 22, 54, 63 & \textbf{95.420} & 29.090 & 3.023 & 0.234 & 0.000 \\
skyscraper & 26, 54, 63 & \textbf{81.134} & 16.823 & 1.975 & 0.350 & 0.023 \\
skyscraper & 22, 26, 54, 63 & \textbf{80.589} & 13.093 & 0.872 & 0.015 & 0.000 \\
\hline
\end{tabular}
\end{footnotesize}}
\end{table}

\begin{table}[p]
\caption{Non-target Label Activation Percentages (Non-TLA) for ADE20K and Google Image dataset: Non-t: percentage of non-target label images that activate the neuron(s) associated with the concept being analyzed across various activation thresholds.}
\label{tab:appendix_non_targ_ade_google}
\centering
\resizebox{\columnwidth}{!}{
\begin{footnotesize}
\begin{tabular}{l|rr|rr|rr|rr}
\hline
Concepts & \multicolumn{2}{c}{non-t $>$0} & \multicolumn{2}{c}{non-t $>$20\%} & \multicolumn{2}{c}{non-t $>$40\%} & \multicolumn{2}{c}{non-t $>$60\%}\\
\hline
 & google & ADE20K & google & ADE20K & google & ADE20K & google & ADE20K\\
\hline
buffet & 32.714 & 40.135 & 12.374 & 25.817 & 3.708 & 9.470 & 0.825 & 1.804 \\
building & 43.375 & 11.458 & 12.314 & 5.208 & 2.276 & 1.458 & 0.182 & 0.000 \\
building and dome & 78.185 & 26.170 & 45.133 & 5.893 & 14.643 & 0.867 & 2.639 & 0.000 \\
central\_reservation & 84.973 & 44.893 & 57.993 & 34.343 & 19.734 & 14.927 & 2.913 & 3.816 \\
clamp\_lamp and clamp & 59.504 & 27.273 & 29.229 & 19.170 & 9.000 & 8.300 & 1.652 & 1.976 \\
closet and air\_conditioning & 71.054 & 30.168 & 38.491 & 15.620 & 10.135 & 5.513 & 1.267 & 1.378 \\
cross\_walk & 28.241 & 21.474 & 6.800 & 16.391 & 1.524 & 9.784 & 0.521 & 2.922 \\
edifice and skyscraper & 48.761 & 24.187 & 21.786 & 8.453 & 8.379 & 1.300 & 2.229 & 0.260 \\
faucet and flusher & 78.562 & 56.967 & 37.862 & 30.580 & 12.104 & 11.097 & 1.873 & 1.850 \\
field & 65.333 & 66.161 & 30.207 & 30.043 & 8.183 & 10.412 & 1.656 & 2.386 \\
flusher and soap\_dish & 63.552 & 19.481 & 29.901 & 10.035 & 7.695 & 3.896 & 1.148 & 0.236 \\
footboard and chain & 66.702 & 27.975 & 40.399 & 13.671 & 17.064 & 5.063 & 4.399 & 1.013 \\
hedgerow and hedge & 68.527 & 45.120 & 30.421 & 28.390 & 7.685 & 13.308 & 1.352 & 2.028 \\
lid and soap\_dispenser & 78.571 & 57.512 & 34.989 & 18.427 & 9.052 & 2.817 & 1.485 & 0.352 \\
mountain & 88.516 & 45.144 & 64.169 & 33.725 & 23.112 & 16.115 & 4.326 & 3.842 \\
mountain and bushes & 24.969 & 28.331 & 10.424 & 16.573 & 4.666 & 6.607 & 1.937 & 1.904 \\
night\_table & 56.714 & 30.534 & 27.691 & 15.267 & 7.691 & 5.954 & 1.137 & 1.679 \\
open\_fireplace and coffee\_table & 16.381 & 26.139 & 4.325 & 10.590 & 0.812 & 2.413 & 0.088 & 0.268 \\
pillow & 46.492 & 12.500 & 9.634 & 3.869 & 0.988 & 1.190 & 0.049 & 0.149 \\
potty and flusher & 76.830 & 58.410 & 36.537 & 24.194 & 10.755 & 4.608 & 1.932 & 1.152 \\
road & 44.592 & 8.501 & 17.727 & 6.955 & 3.471 & 4.328 & 0.702 & 0.927 \\
road and automobile & 41.466 & 17.604 & 16.055 & 14.497 & 3.301 & 8.728 & 0.701 & 2.811 \\
road and car & 48.571 & 14.815 & 25.373 & 11.704 & 8.399 & 6.074 & 3.261 & 1.333 \\
route & 45.089 & 12.349 & 18.937 & 10.241 & 4.809 & 5.723 & 1.169 & 1.807 \\
route and car & 47.408 & 17.073 & 18.871 & 14.204 & 4.081 & 7.461 & 1.416 & 2.152 \\
shower\_stall and cistern & 53.186 & 25.982 & 24.788 & 9.700 & 8.485 & 4.965 & 1.372 & 1.039 \\
Shower\_stall and screen\_door & 31.747 & 24.910 & 12.876 & 14.320 & 4.121 & 5.897 & 1.026 & 1.203 \\
skyscraper & 13.093 & 3.009 & 0.872 & 0.463 & 0.015 & 0.231 & 0.000 & 0.116 \\
slope & 64.503 & 66.520 & 29.976 & 29.967 & 6.894 & 9.879 & 1.200 & 1.976 \\
tap and crapper & 70.606 & 62.225 & 36.839 & 12.861 & 13.696 & 4.890 & 2.511 & 0.611 \\
tap and shower\_screen & 72.584 & 62.621 & 32.574 & 13.180 & 7.836 & 4.733 & 0.860 & 0.607 \\
teapot and saucepan & 47.984 & 23.632 & 18.577 & 11.176 & 4.367 & 6.519 & 0.845 & 1.281 \\
wardrobe and air\_conditioning & 65.034 & 30.525 & 31.795 & 16.160 & 6.958 & 5.525 & 1.145 & 0.967\\

\hline
\end{tabular}
\end{footnotesize}}
\end{table}

\begin{table}[p]
\caption{Statistical Evaluation for \textit{confirmed} concepts (concepts getting $p$-value $<$0.05 for MWU): Non-t: percentage of non-target label images activating the associated neuron(s) analyzed across various activation thresholds.}
\label{tab:appendix_stats_eval}
\centering
\resizebox{.70\columnwidth}{!}{
\begin{footnotesize}
\begin{tabular}{l|rr|r}
\hline
Concepts & Google & ADE20K & p-values\\
\hline
\hline
 & \multicolumn{2}{c}{\textbf{non-t $>$0}} & \\
\hline
building & 43.37468 & 11.45833 & 0.018471 \\
building and dome & 78.185 & 26.16984 & 6.06E-05 \\
central\_reservation & 84.97336 & 44.89338 & 1.75E-66 \\
closet and air\_conditioning & 71.05416 & 30.16845 & 0.009373 \\
edifice and skyscraper & 48.76092 & 24.18726 & 0.016058 \\
faucet and flusher & 78.562 & 56.96671 & 9.19E-07 \\
footboard and chain & 66.702 & 27.97468 & 0.000284 \\
lid and soap\_dispenser & 78.57143 & 57.51174 & 0.00218 \\
pillow & 46.49232 & 12.5 & 4.21E-23 \\
potty and flusher & 76.82974 & 58.41014 & 1.39E-07 \\
shower\_stall and cistern & 53.1865 & 25.98152 & 0.016657 \\
tap and crapper & 70.60579 & 62.22494 & 6.17E-08 \\
tap and shower\_screen & 72.584 & 62.62136 & 0.007024 \\
\hline
\multicolumn{3}{c}{\textbf{Wilcoxon signed rank test (non-t $>$0)}} & \textbf{0.0001221}\\
\hline
\hline
 & \multicolumn{2}{c}{\textbf{non-t $>$20 \%}} & \\
\hline
 building & 12.31365 & 5.208333 & 1.72E-17 \\
building and dome & 45.13343 & 5.892548 & 1.37E-23 \\
clamp\_lamp and clamp & 29.2287 & 19.16996 & 1.57E-07 \\
closet and air\_conditioning & 38.4913 & 15.62021 & 0.000287 \\
edifice and skyscraper & 21.78641 & 8.452536 & 5.80E-17 \\
faucet and flusher & 37.86209 & 30.57953 & 1.80E-15 \\
lid and soap\_dispenser & 34.98939 & 18.42723 & 2.74E-15 \\
mountain and bushes & 10.42437 & 16.57335 & 3.25E-06 \\
pillow & 9.634389 & 3.869048 & 3.49E-49 \\
potty and flusher & 36.53659 & 24.19355 & 3.69E-18 \\
Shower\_stall and screen\_door & 12.87584 & 14.3201 & 0.035051 \\
skyscraper & 0.872071 & 0.462963 & 1.99E-05 \\
tap and crapper & 36.83933 & 12.86064 & 0.000114 \\
tap and shower\_screen & 32.5745 & 13.17961 & 3.22E-14 \\
wardrobe and air\_conditioning & 31.79496 & 16.16022 & 2.18E-11\\
\hline
\multicolumn{3}{c}{\textbf{Wilcoxon signed rank test (non-t $>$ 20\%)}} & \textbf{0.0004272}\\
\hline
\hline
 & \multicolumn{2}{c}{\textbf{non-t $>$40 \%}} & \\
\hline
building & 2.27609 & 1.458333 & 3.16E-19 \\
building and dome & 14.64338 & 0.866551 & 6.28E-20 \\
central\_reservation & 19.73357 & 14.92705 & 1.18E-05 \\
clamp\_lamp and clamp & 9.000096 & 8.300395 & 2.79E-31 \\
closet and air\_conditioning & 10.1354 & 5.513017 & 6.38E-09 \\
cross\_walk & 1.52392 & 9.78399 & 0.000572 \\
edifice and skyscraper & 8.37939 & 1.30039 & 5.06E-17 \\
faucet and flusher & 12.10377 & 11.09741 & 2.90E-24 \\
field & 8.183384 & 10.41215 & 3.82E-05 \\
flusher and soap\_dish & 7.695067 & 3.896104 & 4.26E-08 \\
lid and soap\_dispenser & 9.052334 & 2.816901 & 2.04E-19 \\
mountain and bushes & 4.666314 & 6.606943 & 1.28E-12 \\
pillow & 0.988239 & 1.190476 & 1.37E-23 \\
potty and flusher & 10.75519 & 4.608295 & 1.97E-09 \\
road & 3.471037 & 4.327666 & 0.033105 \\
road and car & 8.399088 & 6.074074 & 0.009958 \\
Shower\_stall and screen\_door & 4.120976 & 5.89651 & 1.13E-07 \\
skyscraper & 0.015367 & 0.231481 & 2.47E-30 \\
slope & 6.893903 & 9.879254 & 1.14E-07 \\
tap and shower\_screen & 7.835857 & 4.73301 & 2.05E-12 \\
wardrobe and air\_conditioning & 6.9579 & 5.524862 & 1.70E-19\\
\hline
\multicolumn{3}{c}{\textbf{Wilcoxon signed rank test (non-t $>$ 40\%)}} & \textbf{0.0479}\\
\hline
\hline
\end{tabular}
\end{footnotesize}}
\end{table}

\begin{table}[p]
\centering
\resizebox{.70\columnwidth}{!}{
\begin{footnotesize}
\begin{tabular}{l|rr|r}
\hline
 & \multicolumn{2}{c}{\textbf{non-t $>$ 60\%}} & \\
\hline
building & 0.182087 & 0 & 1.08E-07 \\
building and dome & 2.639495 & 0 & 5.70E-10 \\
central\_reservation & 2.912966 & 3.815937 & 1.50E-07 \\
clamp\_lamp and clamp & 1.652099 & 1.976285 & 4.24E-19 \\
closet and air\_conditioning & 1.266925 & 1.378254 & 2.50E-07 \\
cross\_walk & 0.520833 & 2.92249 & 0.000171 \\
edifice and skyscraper & 2.228561 & 0.260078 & 4.80E-07 \\
faucet and flusher & 1.872623 & 1.849568 & 0.008524 \\
field & 1.655819 & 2.386117 & 1.43E-09 \\
flusher and soap\_dish & 1.147982 & 0.236128 & 3.03E-13 \\
lid and soap\_dispenser & 1.485149 & 0.352113 & 3.10E-07 \\
mountain and bushes & 1.936961 & 1.903695 & 9.96E-12 \\
pillow & 0.048848 & 0.14881 & 1.04E-09 \\
potty and flusher & 1.931664 & 1.152074 & 0.010232 \\
road & 0.701794 & 0.927357 & 0.000445 \\
road and car & 3.261441 & 1.333333 & 3.79E-05 \\
route and car & 1.415601 & 2.15208 & 0.000137 \\
shower\_stall and cistern & 1.372089 & 1.039261 & 0.031085 \\
Shower\_stall and screen\_door & 1.025822 & 1.203369 & 9.36E-11 \\
skyscraper & 0 & 0.115741 & 6.15E-26 \\
slope & 1.200192 & 1.975851 & 2.39E-10 \\
tap and shower\_screen & 0.859795 & 0.606796 & 3.67E-08 \\
wardrobe and air\_conditioning & 1.144971 & 0.966851 & 1.52E-14\\
\hline
\multicolumn{3}{c}{\textbf{Wilcoxon signed rank test (non-t $>$ 60\%)}} & \textbf{0.05803}\\
\hline
\end{tabular}
\end{footnotesize}}
\end{table}

\end{document}